# Effective injury forecasting in soccer with GPS training data and machine learning

Short title: Injury forecasting in soccer

Alessio Rossi[1*], Luca Pappalardo[1,2], Paolo Cintia[2], F Marcello Iaia[3], Javier Fernàndez[4], Daniel Medina[5]

[1] Department of Computer Science, University of Pisa, Pisa, Italy

[2] ISTI, National Research Council, Pisa, Italy

[3] Department of Biomedical Science for Health, University of Milan, Milan, Italy

[4] Sports Science and Health Department, FC Barcelona, Barcelona Spain

5 Athletic care department, Philadelphia 76ers, Philadelphia, USA

* Corresponding author

E-mail: alessio.rossi2@gmail.com

# Abstract

Injuries have a great impact on professional soccer, due to their large influence on team performance and the considerable costs of rehabilitation for players. Existing studies in the literature provide just a preliminary understanding of which factors mostly affect injury risk, while an evaluation of the potential of statistical models in forecasting injuries is still missing. In this paper, we propose a multi-dimensional approach to injury forecasting in professional soccer that is based on GPS measurements and machine learning. By using GPS tracking technology, we collect data describing the training workload of players in a professional soccer club during a season. We then construct an injury forecaster and show that it is both accurate and interpretable by providing a set of case studies of interest to soccer practitioners. Our approach opens a novel perspective on injury prevention, providing a set of simple and practical rules for evaluating and interpreting the complex relations between injury risk and training performance in professional soccer.

# Introduction

Injuries of professional athletes have a great impact on the sports industry, due to their influence on the mental state of the individuals and the performance of a team [1, 2]. Furthermore, the cost associated with a player's recovery and rehabilitation is often considerable, both in terms of medical care and missed earnings deriving from the popularity of the player himself [3]. Recent research demonstrates that injuries in Spain cause about 16% of season absence by professional soccer players, corresponding to a cost of around 188 million euros per season [4]. It is not surprising, hence, that injury forecasting is attracting a growing interest from researchers, managers, and coaches, who are interested in intervening with appropriate actions to reduce the likelihood of injuries of their players.

Historically, academic work on injury forecasting has been deterred by the limited availability of data describing the physical activity of players. Nowadays, the Internet of Things have the potential to change rapidly this scenario thanks to Electronic Performance and Tracking Systems (EPTS), new tracking technologies that provide high-fidelity data streams extracted from every training and game session [5, 6]. These data depict in detail the movements of players on the playing field [5, 6] and have been used for many purposes, from identifying training patterns [7] to automatic tactical analysis [5, 8, 9]. Despite this wealth of data, little effort has been put on investigating injury forecasting in professional soccer so far [10, 11, 12]. State-of-the-art approaches provide just a preliminary understanding of which variables affect the injury risk, while an evaluation of the potential of statistical models to forecast injuries is still poor. A major limit of existing studies is that they are mono-dimensional, i.e., they use just one variable at a time to estimate injury risk, without fully exploiting the complex patterns underlying the available data.

Professional soccer clubs are interested in practical, usable and interpretable models as a decision making support for coaches and athletic trainers [13]. In this perspective the creation of injury forecasting models poses many challenges. On one hand, injury forecasters must be highly accurate, as models which frequently produce "false alarms" are useless. On the other hand, a "black box" approach (e.g., a deep neural network) is not desirable for practical use since it does not provide any insights about the reason behind the injuries. It goes hence without saying that injury forecasting models must achieve a good tradeoff between accuracy and interpretability.

In this paper, we consider injury prediction as the problem of forecasting that a player will get injured in the next training session or official game, given his recent training workload. We observe that existing mono-dimensional approaches are not effective in practice due to their low precision (< 5%), and we propose a multi-dimensional, easy-to-interpret and fully data-driven approach which forecasts injuries with a better precision (50%); we validate this result by simulating the usage of our forecaster over a season, with new training data available as the season goes by. Our approach is entirely based on automatic data collection through standard GPS sensing technologies and can be a valid supporting tool to the decision making of a soccer club's staff. This is crucial since the decisions of managers and coaches, and hence the success of soccer clubs, also depend on what they measure, how good their measurements are, the quality of predictions and how well these predictions are understood.

## Related work

The relationship between training workload and injury risk has been widely studied in the sports science literature [14, 15, 16, 17, 18]. For example Gabbett et al. [14, 15, 17, 19] investigate the case of rugby and find that a player has a high injury risk when his workloads are increased above certain thresholds. To assess injury risk in cricket, Hulin et al. [20] propose the Acute Chronic Workload Ratio (ACWR), i.e., the ratio between a player's acute workload and his chronic workload. When the acute workload is lower than the chronic workload, cricket players are associated with a low injury risk. In contrast, when the acute/chronic ratio is higher than 2, players have an injury risk from 2 to 4 times higher than the other group of players. Hulin et al. [20] and Ehrmann et al. [11] find that injured players, in both rugby and soccer, show significantly higher physical activity in the week preceding the injury with respect to their seasonal averages.

In skating, Foster et al. [21] measure training workload by the session load, i.e., the product of the perceived exertion and the duration of the training session. When the session load outweighs a skater's ability to fully recover before the next session, the skater suffers from the so-called "overtraining syndrome", a condition that can cause injury [21]. In basketball, Anderson et al. [18] find a strong correlation between injury risk and the so-called monotony, i.e., the ratio between the mean and the standard deviation of the session load recorded in the past 7 days. Moreover, Brink et al. [8] observe that injured young soccer players (age < 18) recorded higher values of monotony in the week preceding the injury than non-injured players.

Venturelli et al. [12] perform several periodic physical tests on young soccer players (age < 18) and find that jump height, body size and the presence of previous injuries are significantly correlated with the probability of thigh strain injury. Talukder et al. [22] create a classifier to predict 19% of the injuries that occurred in NBA. They also show that the most important features for predicting injuries are the average speed, the number of past competitions played, the average distance covered, the number of minutes played to date and the average field goals attempted. An attempt to injury forecasting in soccer has been made by Kampakis [23], although it considers a reduced set of features obtaining an accuracy that is, in the best scenario, not significantly better than random classifiers.

## Material and Method

### Data collection and feature extraction

We set up a study on twenty-six Italian professional male players (age = 26±4 years; height = 179±5 cm; body mass = 78±8 kg) during season 2013/2014. Six central backs, three fullbacks, seven midfielders, eight wingers and two forwards were recruited. Participants gave their written informed consent to participate in the study.

We monitored the physical activity of players during 23 weeks – from January 1st to May 31st, 2014 – using portable 10 Hz GPS devices integrated with a 100Hz 3-D accelerometer, a 3D gyroscope, a 3D digital compass (STATSports Viper). The devices were placed between the players' scapulae through a tight vest. We recorded a total of 931 individual training sessions during the 23 weeks. From the data collected by the devices, we extracted a set of training workload indicators through the software package Viper Version 2.1 provided by STATSports 2014.

The club's medical staff recorded 23 non-contact injuries during the study. According to the UEFA regulations [24], a non-contact injury is defined as any tissue damage sustained by a player that causes absence in physical activities for at least the day after the day of the onset. We observed that 19 out of 23 injuries are associated with players who got injured at least once in the past. In particular, half of the players never get injured during the study, while the others get injured once (seven players), twice (five players) or four times (one player). For every player, we collected information about age, body mass index, height and role on the field. Moreover, for every single training session of a player, we collected information about the play time in the official game before the training session and the number of official games played before the training session.

From the players' GPS data we extract 12 features describing different aspects of the workload in a training session [25]. Two features – Total Distance ($d_{TOT}$) and High Speed Running Distance ($d_{HSR}$) – are kinematic, i.e., they quantify a player's overall movement during a training session. Three features – Metabolic Distance ($d_{MET}$), High Metabolic Load Distance ($d_{HML}$) and High Metabolic

Load Distance per minute ($d_{HML/m}$) – are metabolic, i.e., they quantify the energy expenditure of a player's overall movement during a training session. The remaining seven features – Explosive Distance ($d_{EXP}$), number of accelerations above 2m/s2 ($Acc_2$), number of accelerations above 3m/s2 ($Acc_3$), number of decelerations above 2m/s2 ($Dec_2$), number of decelerations above 3m/s2 ($Dec_3$), Dynamic Stress Load (DSL) and Fatigue Index (FI) – are mechanical features describing a player's overall muscular-scheletrical load during a training session. In addition, we associated a player's training session with feature PI, indicating the number of the player's previous injuries up to that session. Table 1 and S1 Appendix provides the description and some statistics of the workload features extracted from the GPS data, respectively.

**Table 1. Training workload features used in our study.** Description of the training workload features extracted from GPS data and the players' personal features collected during the study. We defined four categories of features: kinematic features (blue), metabolic features (red), mechanical features (green) and personal features (white).

| Feature | Description |
|---|---|
| $d_{TOT}$ | Distance in meters covered during the training session |
| $d_{HSR}$ | Distance in meters covered above 5.5m/s |
| $d_{MET}$ | Distance in meters covered at metabolic power |
| $d_{HML}$ | Distance in meters covered by a player with a Metabolic Power is above 25.5W/Kg |
| $d_{HML/m}$ | Distance in meters covered by a player with a Metabolic Power is above 25.5W/Kg per minute |
| $d_{EXP}$ | Distance in meters covered above 25.5W/Kg and below 19.8Km/h |
| $Acc_2$ | Number of accelerations above 2m/s2 |
| $Acc_3$ | Number of accelerations above 3m/s2 |
| $Dec_2$ | Number of decelerations above 2m/s2 |
| $Dec_3$ | Number of decelerations above 3m/s2 |
| DSL | Total of the weighted impacts of magnitude above 2g. Impacts are collisions and step impacts during running |
| FI | Ratio between DSL and speed intensity |
| Age | age of players |
| BMI | Body Mass Index: ratio between weight (in kg) and the square of height (in meters) |
| Role | Role of the player |
| PI | Number of injuries of the players before each training session |
| Play time | Minutes of play in previous games |
| Games | Number of games played before each training session |

## Multi-dimensional and data-driven injury forecaster

We construct a multi-dimensional model to forecast whether or not a player will get injured based on his recent training workload. The construction of the injury forecaster consists of two phases.

In the first phase (training dataset construction), given a set of features *S*, a training dataset *T* is created where each example refers to a single player's training session and consists of: (i) a vector of features describing both the player's personal features and his recent workload, including the current training session; (ii) an injury label, indicating whether (1) or not (0) the player gets injured in the next game or training session. In the second phase (model construction and validation), a decision tree learner is used to train an injury classifier on the training dataset *T*.

**Phase 1: Training dataset construction**

From the features extracted from GPS data, which are described in Table 1, we construct a training dataset *T* consisting of 55 features and 952 examples (i.e., individual training sessions). S4 Appendix provides an example of the construction of *T*. These 55 features are:

- *18 daily features:* the 12 workload features extracted from the GPS data and the 6 personal features described in Table 1.
- *12 EWMA features:* 12 features computed as the Exponential Weighted Moving Average (EWMA) of the 12 workload features in Table 1. The EWMA decreases exponentially the weights of the values according to their recency, i.e., the more recent a value, the more it is weighted in an exponential function according to a decay α = 2/(span+1). In our experiments we consider a span equal to six (see S5 Appendix).
- *12 ACWR features:* 12 features consisting of the ACWR of the 12 workload features in Table 1. Given a feature, the ACWR of a player is the ratio between (i) the player's acute workload, computed as the average of the values of the feature in the last 6 days; (ii) the player's chronic workload, computed as the average of the values of the feature in the last 27 days [26].
- *12 MSWR features:* 12 features consisting of the monotony of the 12 workload features in Table 1. Given a feature, the monotony of a player is the ratio between the mean and the standard deviation of the values of the feature in the last week [3, 10, 18].

- *1 previous injury feature:* to take into account both the number of a player's previous injuries and their distance to the current training session we compute feature $PI^{(WF)}$, the EWMA of feature PI computed with a span equal to six. $PI^{(WF)}$ reflects the distance between the current training session and the training session when the player returned to regular training after an injury. $PI^{(WF)} = 0$ indicates that the player never got injured in the past; $PI^{(WF)} > 0$ indicates that he got injured at least once in the past; $PI^{(EWMA)} > 1$ indicate that he got injured more than once in the past (see S6 Appendix).

We select 30% of $T$ and obtain $T^{TRAIN}$ (step 1 and 2 in Fig 1) to perform a feature selection process to determine the most relevant features for classification using Recursive Feature Elimination with Cross-Validation (RFECV; we use the publicly available Python package scikit-learn to perform RFECV and to train and validate the decision tree – http://scikit-learn.org/) [27]. In RFECV, the subset of features producing the maximum score on the validation data is considered to be the best feature subset [27]. The feature selection process is aimed at reducing the dimensionality of the feature space and hence the risk of overfitting, and allowing for an easier interpretation of the resulting machine learning model, due to the lower number of features [28].

The class distribution in training dataset $T^{TRAIN}$ is highly unbalanced since we have 279 non-injury examples and just 7 injury examples. To adjust this imbalance we oversample the minority class in $T^{TRAIN}$ by using the adaptive synthetic sampling approach (ADASYN; We use the ADASYN function provided by the publicly available Python package imblearn – http://scikit-learn.org/imbalanced-learn) [29]. The ADASYN algorithm generates examples of the minority class to equalize the distribution of classes, hence reducing the learning bias (See S7 Appendix). Finally, we use $T^{TRAIN}$ to detect the best hyper parameters of a decision tree classifier DT (Step 2 in Fig 1).

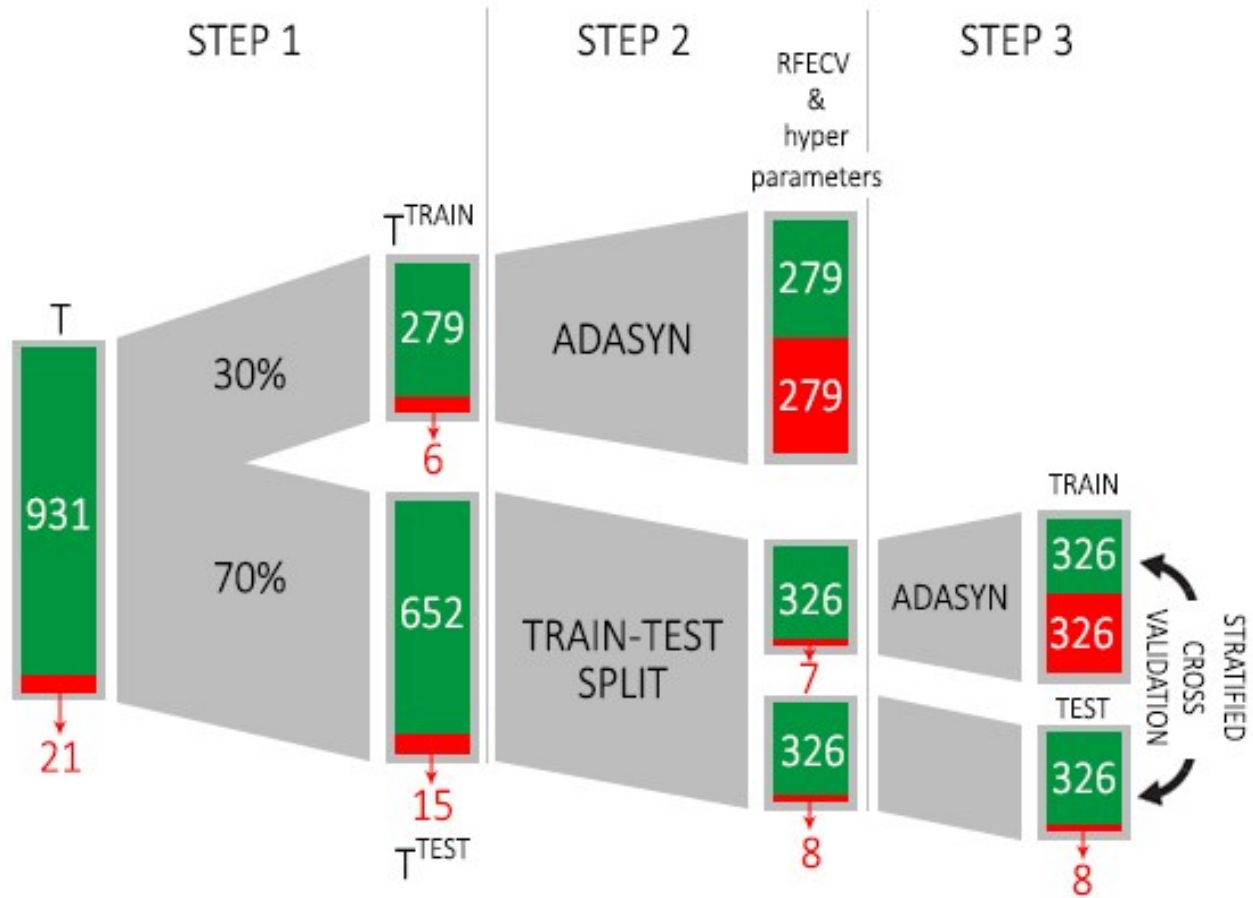

**Fig 1: Construction of the training dataset and the forecasting model.** In step 1 we split the dataset into two parts: $T^{TRAIN}$ (30% of T) and $T^{TEST}$ (70% of T). We then oversample the minority class in $T^{TRAIN}$ by using ADASYN, select the most important features and fit the hyper parameters (Step 2). We then split $T^{TEST}$ into two folds in order to perform a stratified cross validation (step 3).

## Phase 2: Model construction and validation

We then split $T^{TEST}$ into two folds, $f_1$ and $f_2$, in order to perform a stratified cross validation (step 3 in Fig 1; we use only two folds in order to not excessively reduce the minority class size). In this step, we oversample fold $f_1$ by using ADASYN and test DT on the other fold $f_2$ (which is not oversampled). For cross validation purposes, we perform again step 3 inverting $f_1$ and $f_2$. The goodness of the forecasting model is evaluated by four metrics (i.e., precision, recall, F1-score and AUC) described in S8 Appendix. Note that, for injury forecasting purposes, we are interested in achieving high values of precision and recall on class 1 (injury). Let us assume that a coach makes a decision about whether or not to "stop" a player based on the suggestion of the injury forecaster, i.e., the player skips next training session or game every time the forecaster's prediction associated with the player's current training session is 1 (injury). In this scenario, the forecaster's precision indicates how much we can trust the predictions: the higher the precision, the more a classifier's predictions are reliable, i.e., the probability that the player will actually get injured is high. Trusting an injury forecaster with low precision is risky as it means producing many false positives (i.e., false alarms) and frequently stopping players unnecessarily, a condition clubs want to avoid especially for the key players. The recall indicates the fraction of injuries the forecaster detects over the total number of injuries: the higher the recall the more injuries the forecaster can detect. An injury forecaster with low recall detects just a small fraction of the injuries, meaning that many players will attend next training session or game and actually get injured. Trusting a forecaster with a low recall is risky as it would misclassify many actual injuries as non-injuries.

We repeated the entire injury prediction approach (i.e., all the three steps in Fig 1) 10,000 times in order to assess its stability with respect to the choice of the injury examples in the two folds. For the sake of comparison, we implemented other injury forecasters based on the ACWR and the monotony (or MSWR) techniques, which are among the two most used techniques for injury risk estimation and prediction in professional soccer (see S2 Appendix and S3 Appendix for details). Moreover, we compare our injury forecaster with four baselines. Baseline $B_1$ randomly assigns a class to an example by respecting the distribution of classes. Baseline $B_2$ always assigns the non-injury class, while baseline $B_3$ always assigns the injury class. Baseline $B_4$ is a classifier which assigns class 1 (injury) if $PI^{(EWMA)} > 0$, and 0 (no injury) otherwise. We also compare DT with a Random Forest classifier (RF) and a Logit classifier (LR).

# Results

Table 2 compares the performance of DT with the performance of RF, LR, the ACWR and MSWR forecasters, and the four baselines. The results in Table 2 refer to the mean and the standard deviation of the evaluation metrics over 10,000 cross validation tasks. We find that DT has recall=0.80±0.07 and precision=0.50±0.11 on the injury class, meaning that the decision tree can predict almost all the injuries (80%) and that it correctly labels a training session as an injury in 50% of the cases. This is a significant improvement with respect to both the baselines $B_1,...,B_4$, for which the maximum precision is about 6%, and the ACWR- and MSWR-based injury forecasters, for which the maximum precision is lower than 4%. RF has better recall but worse precision (recall=0.87±0.05, precision=0.41±0.08) that DT, while LR has much lower performance than the decision tree (Table 2). These results show that, typically, DT drastically reduces false alarms and hence scenarios where players are "stopped" unnecessarily before next game or training session. On the one hand, the distributions of the forecasters' performances over the 10,000 tests indicate that the quality of the injury forecasting strongly depends on the type of injuries in the training set, which in turn depends on the different training and test split made in each trial (Fig 2). On the other hand, the higher performance detected by DT, compared to several baselines and the ACWR- and MSWR-based injury forecasters, shows that our approach outperforms state-of-the-art approaches and achieve good results in forecasting injuries. The results for DT without ADASYN and the oversampling process are presented in S9 Appendix.

As a further test of the forecasting potential of our approach we investigate the benefit of using our multi-dimensional injury forecaster in a real-world injury prevention scenario, where we assume that a club equips with appropriate GPS sensor technologies and starts recording training workload data since the first training session of the season (in other words, no data are available to the club before the beginning of the season). Assuming that we train the injury forecaster with new data every week, how many injuries the club can actually prevent throughout the season?

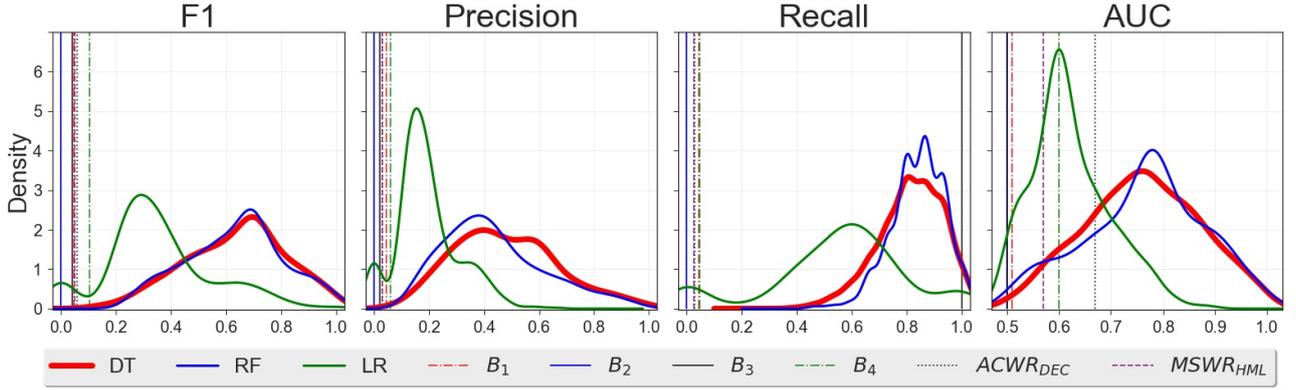

**Fig 2. Classifiers performances.** Distributions of the classifiers - DT, LR and RF - performances obtained testing the algorithms 10,000 times. This figure shows the performance of the baselines and the ACWR- and MSWR-based injury forecasters as well.

**Table 2: Performance of DT compared to RF, LR, the four baselines and the ACWR- and MSWR-based forecasters.** For each forecaster we report precision, recall and F1 on the two classes and the overall AUC.

| | | precision | recall | F1 | AUC |
|---|---|---|---|---|---|
| DT | NI | 0.96±0.05 | 0.87±0.09 | 0.91±0.04 | 0.76±0.12 |
| | I | 0.50±0.11 | 0.80±0.07 | 0.64±0.10 | |
| RF | NI | 0.94±0.06 | 0.90±0.08 | 0.93±0.07 | 0.78±0.15 |
| | I | 0.41±0.08 | 0.87±0.05 | 0.65±0.08 | |
| LR | NI | 0.69±0.11 | 0.61±0.15 | 0.65±0.13 | 0.60±0.03 |
| | I | 0.18±0.03 | 0.60±0.08 | 0.31±0.06 | |
| B4 | NI | 0.98 | 0.77 | 0.86 | 0.60 |
| | I | 0.04 | 0.43 | 0.07 | |
| B1 | NI | 0.98 | 0.98 | 0.98 | 0.51 |
| | I | 0.06 | 0.05 | 0.05 | |
| B2 | NI | 0.98 | 1.00 | 0.99 | 0.50 |
| | I | 0.00 | 0.00 | 0.00 | |
| B3 | NI | 0.00 | 0.00 | 0.00 | 0.50 |
| | I | 0.02 | 1.00 | 0.04 | |
| $C^{(ACWR)}_{DEC}$ | NI | 1.00 | 0.43 | 0.60 | 0.67 |
| | I | 0.04 | 0.91 | 0.07 | |
| $C^{(MSWR)}_{HML}$ | NI | 0.98 | 0.80 | 0.88 | 0.57 |
| | I | 0.04 | 0.33 | 0.07 | |

To answer this question we group the training sessions by week and proceed from the least recent to the most recent week. At training week $w_i$ we first construct the dataset $T_i$ consisting of all the training examples collected up to week $i$, oversampling the injury examples through ADASYN and reducing the feature space through RFECV. Then, we use $T_i$ to train $DT_i$, $RF_i$, $LR_i$, $B_{1,i}, ..., B_{4,i}$, the ACWR- and MSWR-based forecasters and try to predict the injuries in week $w_{i+1}$. At week $i$, we evaluate the accuracy of our approach by the cumulative F1-score, i.e., the F1-score computed by considering all the predictions made up to week $i$ by the models $DT_6, ..., DT_i$. Due to the initial scarcity of data, we start the forecasting task from week $w_6$.

Fig 3 and S7 Table show the evolution of the cumulative F1-score and the feature extracted by RFECV as the season goes by, respectively. We find that in the first weeks DT has a poor predictive performance and misses many injuries (the black crosses in Fig 3). The predictive ability of DT improves significantly throughout the season: as more and more training and injury examples are collected, the forecasting model predicts most of the injuries in the second half of the season (the red crosses in Fig 3). We find that DT is the one performing the best, outperforming all the other models from week $w_{14}$. In particular, DT detects 9 injuries out of 14 from $w_6$ to the end of the season, resulting in F1-score=0.60 and precision=0.56. After an initial period of data collection, the injury forecaster becomes a useful tool to prevent the injuries of players and, by extracting the rules from the decision tree as we show in the next section, to understand the reasons behind the forecasted injuries as well as the injuries that are not detected by the model.

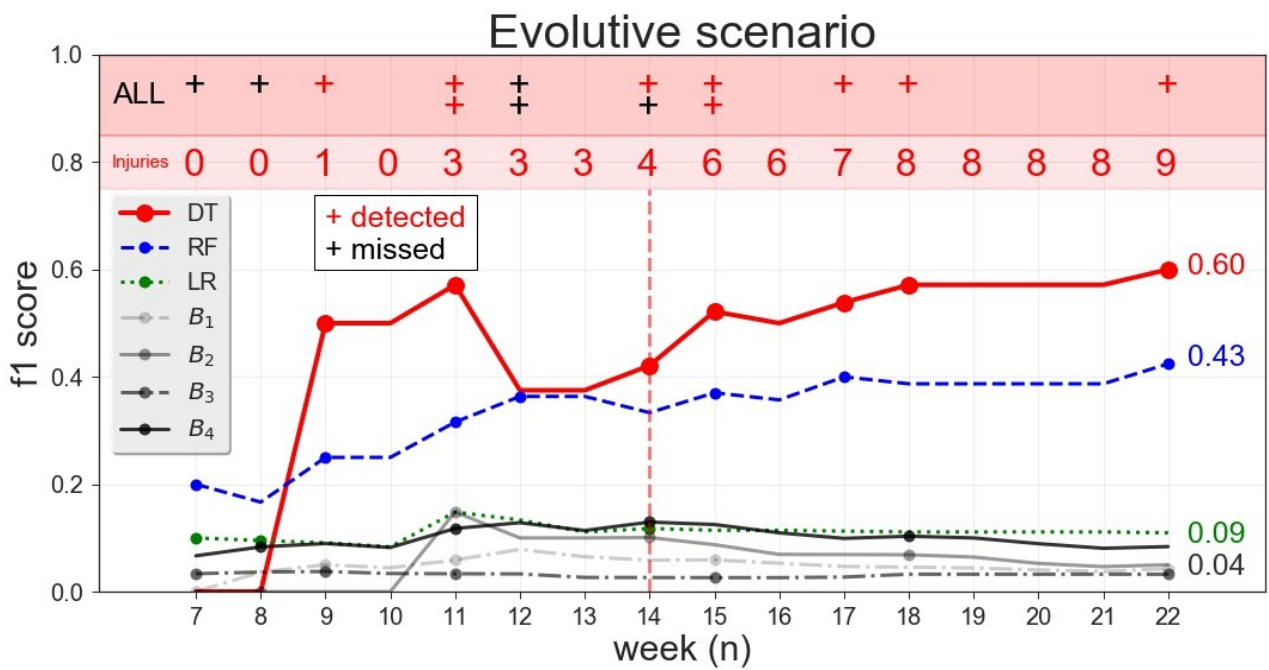

**Fig 3: Performance of forecasters in the evolutive scenario.** As the season goes by, we plot week by week the cumulative F1-score of the forecasters DT, RF, LR, $B_1,...,B_4$ trained on the data collected up to that week. Black crosses indicate injuries that not detect by DT, red crosses indicate injures correctly predicted by DT. For every week $i$ we highlight in red the number of injuries detected by DT up to week $i$.

**Interpretation of the injury forecaster**

A set of simple rules can be extracted from DT build on $w_{21}$, allowing for the investigation of the reasons behind the observed injuries. These rules can be seen as a short handbook for coaches and athletic trainers, which can consult it to modify the training schedule and improve the players' fitness.

Fig 4b visualizes DT highlighting two types of node: decision nodes (black boxes) and leaf nodes (green or red boxes). Each decision node has two branches each indicating the next node to select depending on the range of values of the feature associated with the decision node. A leaf node represents the final prediction based on a player's individual training session. There are two possible final decisions: Injury (red boxes) indicates that the player will get injured in next game or training session; or No-Injury (green boxes) otherwise. Given a feature vector describing a player's training session, the prediction associated with it is obtained by following the path from the root of the tree down to a leaf node, through the decision nodes. Fig 4 shows the rules and the tree extracted from the DT built until $w_{21}$. At the end of the season, the RFECV process selects just 3 features out of 55: $PI^{(EWMA)}$, $d_{HSR}^{(EWMA)}$ and $d_{TOT}^{(MSWR)}$. The importances of these features in DT, computed as the mean decrease in Gini coefficient, are 0.71, 0.23 and 0.06, respectively [30].

As a practical example of application of these rules, let us consider a player's training session with $PI^{(EWMA)} = 0.28$, $d_{HSR}^{(EWMA)} = 126.58$ and $d_{TOT}^{(MSWR)} = 1.66$, associated with an injury. This example is associated with rule 2 (Fig 4a), corresponding to the following decision path:

$$d_{HSR}^{(EWMA)} > 112.35 \rightarrow d_{TOT}^{(MSWR)} \leq 1.78 \rightarrow PI^{(EWMA)} > 0.03 \rightarrow PI^{(EWMA)} \leq 0.68 \rightarrow \textbf{INJURY}$$

From the rules in Fig 4a we summarize three main injury scenarios in DT:

1) a previous injury can lead to a new injury when a player has a $HSR^{(EWMA)}$ (high speed running distance) lower than 112.35 (rule 1 in Fig 4a). This rule describes 42% of the injuries in the dataset and it is correct in 100% of the cases.

2) a previous injury can lead to a new injury when a player has a HSR$^{(EWMA)}$ higher than 112.35 and a D$_{tot}^{(MSWR)}$ (total distance Monotony) three times lower than 1.78 (rule 2 in Fig 4a). This rule describes 30% of the injuries and has an accuracy of 100%.

3) a previous injury can lead a new injury when a player has a HSR$^{(EWMA)}$ higher than 112.35 and a D$_{tot}^{(MSWR)}$ two and half times higher than the player's average (rules 3 and 4 in Fig 4a). These rules have a cumulative frequency of 28% and a mean accuracy of 75±5%.

These scenarios suggest that coaches and athletic trainers must take care of the total distance and the distance at high speed running performed by the players who recently returned to play after an injury.

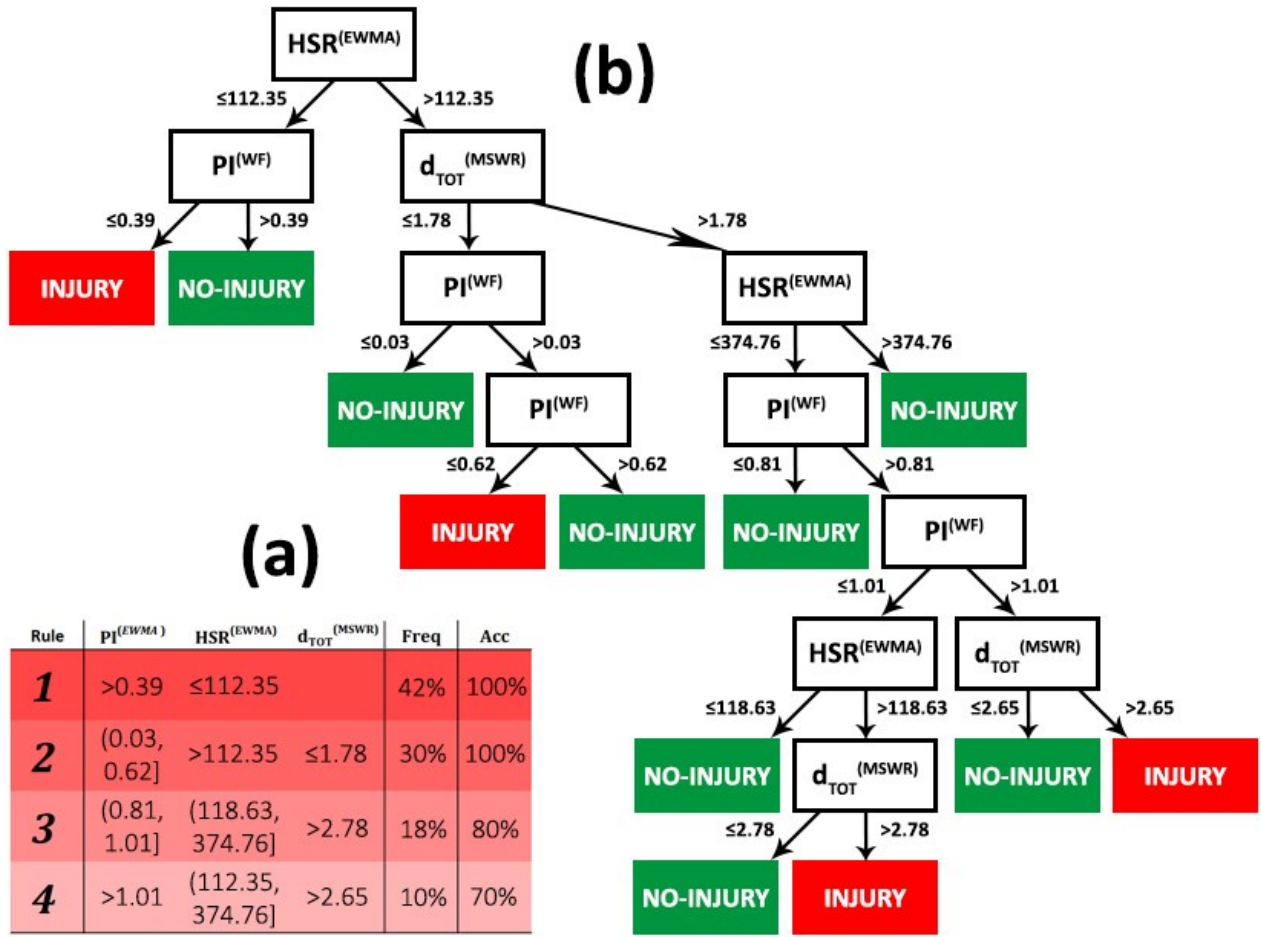

**Fig 4: Interpretation of the multi-dimensional injury forecaster. (a)** The six injury rules extracted from DT. For each rule we show the range of values of every feature, its frequency (Freq) and accuracy (Acc). **(b)** A schematic visualization of decision tree. Black boxes are decision nodes, green boxes are leaf nodes for class No-Injury, red boxes are leaf nodes for class Injury.

# Discussion

Our experiments produce three remarkable results. First, DT can detect around 80% of the injuries with about 50% precision, far better than the baselines and state-of-the-art injury risk estimation techniques (see Table 2). The decision tree's false positive rate is small, indicating that it reduces the "false alarms", i.e., situations where the classifier is wrong in predicting that an injury will happen. In professional soccer, false alarms are deprecable because the scarcity of players can negatively affect the performance of a team [2]. Our model also produces a moderate false negative rate, meaning that situations where a player that will get injured is classified as out of risk are infrequent.

Second remarkable results is that, in a real-world scenario of injury prevention where a club starts collecting the data for the first time and re-train the injury forecaster as the season goes by, the injury forecaster results in a cumulative F1-score=0.60 on the injury class (Fig 3), much better than the baselines, RF and LR (Table 2). Throughout the season, the usage of the forecasting model allows for the prevention of more than half of the injuries. The forecasting ability of DT is affected by the initial period where data are scarce. This suggests that an initial period of data collection is needed in order to gather the adequate amount of data, and only then a reliable forecasting model can be trained on the collected data. The length of the data collection period depends on the club's needs and strategy, including the frequency of training sessions and games, the frequency of injuries, the number of available players and the tolerated level of false alarms. Regarding this aspect, in our dataset, we observe that the performance of the classifiers stabilizes after 14 weeks of data collection (see Fig 3).

Third, in the evolutive scenario the features selected change as the season goes by (see S7 Table). This is probably due to the initial scarcity of data and to the type of injuries that have occurred up that a given moment. We observed that the just 3 out of 55 features are selected by the feature selection ($PI^{(EWMA)}$, $d_{HSR}^{(EWMA)}$ and $d_{TOT}^{(MSWR)}$) after 14 weeks of data collection, and that these set of features remains stable for all subsequent weeks. Feature $PI^{(EWMA)}$, the most important among the three and the only feature that is always selected as the season goes by (see S7 Table), reflects the temporal distance between a player's current training session and his coming back to regular training after a previous injury. Less than half of the injuries detected by DT in the evolutive scenario happened immediately after the coming back to regular training of injured player. Furthermore, 60% of the injuries detected by DT happened long after a previous injury and are characterized by specific values of $d_{HSR}^{(EWMA)}$ and $d_{TOT}^{(MSWR)}$, which indicate that the a player's kinematic variability affects his injury risk. It is worth to notice that the single feature $PI^{(EWMA)}$ alone does not provide a significant predictive power, as the baseline $B_4$, which is based on it, has a much lower accuracy than DT. It is hence the combination of the three features which allows us to predict when a player will get injured. Our results suggest that the club should take particular care of the first training sessions of players who come back to regular training after a previous injury, as in this conditions they are more likely to get injured again. In these first days and in the days long after the players return to regular physical activity, the club should control kinematic workloads, which can lead to injuries at specific values as well.

Injuries involve a great economic cost to the club, due to the expensive process of recovery and rehabilitation for the players. Injury prevention can reduce these costs by avoiding the injuries of players, which means improving the team's performance and the player's mental state as well as reducing the seasonal costs of medical care. We estimate that 139 days of absence during the seasons are due to injuries, corresponding to 6% of the working days. We observe that a player returned to regular physical activity within 5 days (i.e., 15 times out of 23 injuries), while only 6 times a player needed more than 5 days to recover. We use a method proposed in the literature [4] to estimate that the minimum total cost related to injuries that in this soccer club is 11,583 euros (139x83 euros = days of absence x minimal legal salary per day) corresponding to 3.81% of the salary cost of the club. If our model was used as the season goes by to stop the players for which an injury is predicted, the club could had been able to prevent 9 injuries out of 14 and save 8,881 euros (107x83 euros = day of absence x minimal legal salary per day), that represents a 77% decrease of injury costs.

# Conclusion

In this paper we proposed a multi-dimensional approach to injury forecasting in soccer, fully based on automatically collected GPS data and machine learning. As we showed, our injury forecaster provides a good trade-off between accuracy and interpretability, reducing the number of false alarms with respect to state-of-the-art approaches and at the same time providing a simple handbook of rules to understand the reasons behind the observed injuries. We showed that the forecaster can be profitably used early in the season, and that it allows the club to save a considerable part of the seasonal injury-related costs. Our approach opens a novel perspective on injury prevention, providing a methodology for evaluating and interpreting the complex relations between injury risk and training performance in professional soccer.

Our work can be extended in many directions. First, we can include performance features extracted from official games, where the player is exposed to the highest physical and psychological stress. Second, we can investigate the "transferability" of our approach from a club to another, i.e., if a forecaster trained on a set of players can be successfully applied to a distinct set of players, not used during the training process. In this case, it would be possible to exploit collective information to train a more powerful forecaster which includes training examples from different players, clubs, and leagues. Third, if data covering several seasons of a player's activity are available, a distinct forecaster can be trained for each player by combining GPS data with other types of health data, such as heart rate, ventilation, and lactate.

**Acknowledgements.** This work is partially supported by the European Community's H2020 Program under the funding scheme "INFRAIA-1-2014-2015: Research Infrastructures" grant agreement 654024, www.sobigdata.eu, "SoBigData".

# SUPPORTING INFORMATION

**S1 Appendix. Descriptive statistics of the workload features.**

S1 Table shows the average (AVG) and the standard deviation (SD) of the distributions of the 12 training workload features considered in our study. We assess the normality of the distributions by using the Shapiro-Wilks' Normality test (SW) and observe that none of them is normally distributed (see S1 Table). Indeed, by a visual inspection of the distributions, we observe that they tend to be bimodal and right skewed (S1 Fig).

**S2 Appendix. The ACWR method**

The Acute Chronic Workload ratio (ACWR), defined as the ratio between a player's acute workload and his chronic workload [14, 20, 26], is one of the most used technique for injury risk estimation in professional soccer [31]. A player's acute and chronic workloads are estimated by the exponential weighted moving average of a single workload feature *h* in the previous 7 days and 28 days, respectively. The player's ACWR is then used to estimate his injury likelihood [26].

We reproduce the ACWR methodology for each of the 12 workload features used in our study, using the ACWR groups suggested by Murray et al. [26]. They compute the ACWR for a set of workload features and categorize the players' training sessions with five groups: (1) ACWR < 0.49 (very low); (2) ACWR [0.50, 0.99] (low); (3) ACWR [1.00, 1.49] (moderate); (4) ACWR [1.50, 1.99] (high); (5) ACWR > 2.00 (very high). Then, the injury likelihood (IL) is estimated in every ACWR group as the ratio between the number of players who get injured after the training session assigned to that ACWR group and the number of players who do not. Murray et al. [26] observe that players whose training sessions result in ACWR > 2 have a higher injury risk than the players in the other groups (i.e., a high IL). In contrast with the literature, we do not find any individual training session resulting in ACWR > 2 (see S2 Fig), while we observe that players whose individual training sessions result in ACWR < 1 have the highest injury risk (S2 Fig).

Additionally, we explore the usability in practice of the ACWR method by constructing injury forecasting models based on the ACWR method. In particular, given a player's training session, a predictive model $C_h^{(ACWR)}$ predicts whether or not the player will get injured during next game or training session based on the value of workload feature $h$. If considering feature $h$ the individual training session results in ACWR < 1, $C_h^{(ACWR)}$ forecasts an injury (class 1) otherwise it forecasts a non-injury (class 0). We find that $C_h^{(ACWR)}$ has in average a high recall ($0.80 \pm 0.08$) but a very low precision ($0.03 \pm 0.003$), denoting the presence of a high rate of false alarms, as in average the models wrongly predict an injury in 97% of the cases. Moreover, we combine the ACWR forecasters in three ways:

- the predictive model $C_{(vote)}$ predicts a player will get injured if his training session results in ACWR < 1 for the majority of the workload features;
- the predictive model $C_{(all)}$ predicts that a player will get injured if his training session results in ACWR < 1 for all the workload features;
- the predictive model $C_{(one)}$ predicts an injury if ACWR < 1 for at least one workload feature.

S2 Table reports the accuracy of $C_{(vote)}$, $C_{(all)}$, $C_{(one)}$. Only $C_{(vote)}$ achieves a slightly better performance, in terms of precision on the injury class, than the ACWR forecasters based on the single features. We compare the predictors with four baselines. Baseline $B_1$ randomly assigns a class to an example by respecting the distribution of classes. Baseline $B_2$ always assigns the majority class (i.e., class 0, a non-injury), while baseline $B_3$ always assigns the minority class (i.e., class 1, injury). Baseline $B_4$ is a classifier which assigns class 1 (injury) if the exponentially weighted average of variable PI > 0 (see S5 Appendix), and 0 (no injury) otherwise. Although the 15 ACWR forecasters are significantly better than the baseline classifiers in terms of recall on the injury class, our results suggest that a predictor based on ACWR is not usable in practice due to its low precision. In a scenario where a coach or an athletic trainer bases his decisions on the suggestions of the predictors, in the vast majority of the cases he would generate "false alarms" by stopping a player with no risk of injury, which is not a practical solution to injury prevention in professional soccer.

We also replicate the experiment by using ACWR groups defined by the quintiles of the ACWR distribution instead of the pre-defined groups proposed by Murray et al. [26]. S3 Fig shows the injury likelihood (IL) for every ACWR quintile for all the 12 workload features. We observe that the groups with low ACWR are associated to the highest injury risk, substantially confirming the experiments made using predefined ACWR groups. We construct predictors $C_h^{(ACWRq)}$ following the same strategy as above but using quantiles instead of predefined groups. S3 Table visualizes the results of classification, which are similar to those presented in the manuscript for the pre-defined ACWR groups: the predictors are a little usable in practice due to their too low precision.

**S3 Appendix. The MSWR method**

Another widely used method for injury risk estimation is the Mean Standard deviation Workload Ratio (MSWR, or Monotony), defined as the ratio between the mean and the standard deviation of a single player's workload feature $h$ obtained in one week [10, 18, 21]. High MSWR values are generally associated with negative game performance and high injury risk [21].

We investigate the relation between MSWR and injury risk by grouping the individual training sessions into quintiles according to the distribution of the workload features. For every quintile, we compute the corresponding injury likelihood (IL). We observe that high MSWR values are related to high injury risk for the majority of workload features, substantially confirming results observed in the literature by Foster et al. [30] (see S4 Fig).

As done for ACWR, we explore the usability in practice of MSWR by constructing 12 predictive models based on the 12 training workload features. Given a player's training session, every predictive model $C_{MSWR}$ predicts whether or not the player will get injured during next game or training session based on the value of workload feature $h$. If considering feature $h$ the individual training session is associated with the MSWR group with the highest injury risk, model $C_{MSWR}$ predicts an injury (class 1), otherwise it predicts a non-injury (class 0). We find that $C_h^{(MSWR)}$ has in average both a low recall (0.10 ± 0.10) and a low precision (mean is 0.03 ± 0.03, see S4 Table).

Moreover, we construct three combined models – $C_{(vote)}$, $C_{(all)}$ and $C_{(one)}$ – and we observe that they have poor accuracy in detecting the injury class (S4 Table). In particular, the MSWR predictors have predictive power comparable to the ACWR predictors.

**S4 Appendix. Example of the training dataset construction**

Let us consider a toy dataset consisting of a portion of the training sessions of a player $D = \{s_6, s_7, s_8, s_9\}$ where the last session ($s_9$) is associated with an injury, i.e., the player will get injured during training session $s_{10}$. We construct the training dataset $T$ as follows:

(1) We create a new example in dataset $T$ for each training session in $D$, by computing 42 player's workload features. Every example is described by a vector of length 42, $m_i = (h_1, \ldots, h_{42})$. All the four vectors compose matrix $F = (m_1, m_2, m_3, m_4)$;

(2) Since the first three training sessions are not associated with injuries, the first three examples $m_1, m_2, m_3$ have injury label 0. The last example $m_4$ has injury label 1 since it is associated with an injury. Therefore, the labels vector is hence $c = (0, 0, 0, 1)$, indicating that the first three examples are not associated with an injury while the last training session produces an injury. The training dataset based on $D$ and feature set *all* is finally $T = (F, c)$.

**S5 Appendix. Exponential Weighted Moving Average (EWMA)**

To consider the recent training workload of a player, we compute the exponential weighted moving average (ewma) of his most recent training sessions. The ewma decreases exponentially the weights of the values according to their recency [32, 33], i.e., the more recent a value is the more it is weighted in an exponential function according to a decay $\alpha = 2/(\text{span } 1)$. In accordance with the exponential function, the moving average is computed as:

$$EWMA_t = \alpha[x_t - (x_{t-1} + (1-\alpha)^2 x_{t-2} + \ldots + (1-\alpha)^{n-1} x_{t-n})] + x_t$$

We vary span = 1, … , 10 to detect the value leading to the best classification performance. We hence train a decision tree on the feature set all by using every of the ten span values. Fig 10 shows the cross-validated AUC and F1-score of the decision tree $DT^{(RFE)}$ varying the value of span. We observe that a 6 training span is the best predictive window to injury prediction in our dataset (S5 Fig).

**S6 Appendix. Computation of $PI^{(WF)}$**

To take into account the injury history of a player, we compute the EWMA of the number of injuries in previous weeks. $PI^{(WF)}$ reflects the temporal distance between a player's training session and the return of the player to regular training after an injury. $PI^{(WF)} = 0$ represents players who never got injured in the past. $PI^{(WF)} > 0$ represents players who get injured at least once in the past. S5 Table provides specific $PI^{(WF)}$ thresholds in players incurred from 1 to 4 previous injuries. For example, $PI^{(WF)} = 0.50$ reflects a training performed by a player 3 days since his return to regular training after an injury.

**S7 Appendix. Adaptive synthetic sampling approach**

For each new instance of the minority class in $T^{TRAIN}$, ADASYN automatically decides the number of synthetic samples that need to be generated according to a density distribution of the majority class in $T^{TRAIN}$. The dataset resulting from ADASYN provides a balanced representation of the data distribution. Each new example is computed as $x_i+(x_{zi}-x_i)\lambda$ where $(x_{zi}-x_i)$ is the difference vector in a n-dimensional space and $\lambda$ is an arbitrary number between 0 and 1 randomly assigned in each new example [29].

**S8 Appendix. Classifiers metrics assessment**

1. **Precision**: prec = TP/(TP+FP), where TP and FP are true positives and false positives resulting from classification, where we consider the injury class (1) as the positive class. Given a class, precision indicates the fraction of examples that the classifier correctly classifies over the number of all examples the classifier assigns to that class;

2. **Recall**: rec = TP/(TP + FN), where FN are false negatives resulting from classification. Given a class, the recall indicates the ratio of examples of a given class correctly classified by the classifier;

3. **F1-measure**: F1 = 2(prec * rec)/(prec + rec). This measure is the harmonic mean of precision and recall, which in our case coincides with the square of the geometric mean divided by the arithmetic mean;

4. **Area Under the Curve (AUC)**: the probability that a classifier will rank a randomly chosen positive instance higher than a randomly chosen negative one (assuming "positive" ranks higher than "negative"). An AUC close to 1 represents an accurate classification, while an AUC close to 0.5 represents a classification close to randomness.

**S9 Appendix. Predictions results**

*Oversampling without feature selection* - Here we provide the performance of the classifiers trained without the feature selection process (i.e., we use the same approach provide in Figure 1 without the feature selection process shows in the step 2).
We find that $DT^{(ADA)}$ has precision=0.88 and recall=0.92 on the injury class. Although $DT^{(ADA)}$ (i.e., scenario without feature selection) has a performance comparable to $DT^{(ADA+RFE)}$ (i.e., scenario with feature selection), the latter uses just 3 features out of 55, resulting in a decision tree much easier to interpret and understand.

*No-oversampling without feature selection* - Here we provide the performance of the classifiers trained on the unbalanced training dataset T (931 no-injury and 23 injury examples). On this dataset we train $DT^{(T)}$, $RF^{(T)}$ and $LR^{(T)}$. We validate the classifiers with a 3-fold stratified cross-validation strategy: the real dataset is divided into 3 parts or folds and, for each fold, we use the 10% of the target values as test set, and the remaining 90% as training set.

$DT^{(T)}$ has precision=0.70 and recall=0.47 on the injury class. $RF^{(T)}$ provides just a tiny improvement in terms of recall, but not in precision (precision=0.88, recall=0.60), while $LR^{(T)}$ has much lower performance (precision=0.58, recall 0.33) than $DT^{(T)}$.

*No-oversampling with feature selection* - Here we provide the performance of the classifiers trained on the unbalanced training dataset T (931 no-injury and 23 injury examples) on which we perform feature selection by RFECV to determine the most relevant features for classification. We detected that $PI^{(EWMA)}$, $d_{HML}^{(MSWR)}$ and $Dec_2^{(EWMA)}$ are the most important features. Second, on the new training dataset $T^{(RFE)}$ derived from the feature selection, we train $DT^{(RFE)}$, $RF^{(RFE)}$ and $LR^{(RFE)}$. We validate the classifiers with a 3-fold stratified cross-validation strategy: the real dataset is divided into 3 parts or folds and, for each fold, we use the 10% of the target values as test set, and the remaining 90% as training set.

$DT^{(RFE)}$ is able to detect 56% of the injuries with a precision of 74%. $RF^{(RFE)}$ provides just a tiny improvement in terms of recall, but not in precision (precision=0.78, recall 0.58), while $LR^{(ADA)}$ has much lower performance (precision=0.73, recall 0.48) than $DT^{(RFE)}$.

**S1 Table. Descriptive statistics of the 12 training workload features.** We provide three categories of training workload features: kinematic features (blue), metabolic features (red) and mechanical features (green).

|  | AVG | SD | SW |
|---|---|---|---|
| $d_{TOT}$ | 3882.94 | 1633.21 | <0.01 |
| $d_{HSR}$ | 133.22 | 66.41 | <0.01 |
| $d_{MET}$ | 1151.99 | 694.25 | <0.01 |
| $d_{HML}$ | 543.89 | 339.64 | <0.01 |
| $d_{HML/m}$ | 8.70 | 6.09 | <0.01 |
| $d_{EXP}$ | 410.67 | 221.29 | <0.01 |
| $Acc_2$ | 64.26 | 31.72 | <0.01 |
| $Acc_3$ | 16.16 | 10.97 | <0.01 |
| $Dec_2$ | 62.44 | 33.09 | <0.01 |
| $Dec_3$ | 19.14 | 12.78 | <0.01 |
| DSL | 117.98 | 78.52 | <0.01 |
| FI | 0.63 | 0.31 | <0.01 |

SD = Standard Deviation;
SW = Shapiro-Wilks' Normality test.

**S2 Table. Performance of ACWR predictor.** We report precision (prec), recall (rec), F1-score (F1) and Area Under the Curve (AUC) for the injury class and the non- injury class for all the predictors based on ACWR and MSWR. We also provide predictive performance of four baseline predictors $B_1$, $B_2$, $B_3$ and $B_4$.

| ACWR | class | prec | rec | F1 | AUC |
|---|---|---|---|---|---|
| $C_{dTOT}$ | 0 | 0.99 | 0.44 | 0.61 | 0.65 |
| | 1 | 0.03 | 0.86 | 0.06 | |
| $C_{dHSR}$ | 0 | 0.99 | 0.37 | 0.54 | 0.57 |
| | 1 | 0.03 | 0.76 | 0.05 | |
| $C_{dMET}$ | 0 | 0.99 | 0.43 | 0.60 | 0.59 |
| | 1 | 0.03 | 0.76 | 0.06 | |
| $C_{dHML}$ | 0 | 0.99 | 0.43 | 0.60 | 0.60 |
| | 1 | 0.03 | 0.76 | 0.06 | |
| $C_{dHML/m}$ | 0 | 0.99 | 0.39 | 0.56 | 0.60 |
| | 1 | 0.03 | 0.81 | 0.06 | |
| $C_{dEXP}$ | 0 | 1.00 | 0.43 | 0.60 | 0.67 |
| | 1 | 0.04 | 0.91 | 0.07 | |
| $C_{Acc2}$ | 0 | 0.99 | 0.47 | 0.64 | 0.64 |
| | 1 | 0.03 | 0.80 | 0.06 | |
| $C_{Acc3}$ | 0 | 0.99 | 0.45 | 0.64 | 0.58 |
| | 1 | 0.03 | 0.71 | 0.06 | |
| $C_{Dec2}$ | 0 | 0.99 | 0.46 | 0.63 | 0.66 |
| | 1 | 0.04 | 0.86 | 0.07 | |
| $C_{Dec3}$ | 0 | 0.99 | 0.46 | 0.63 | 0.66 |
| | 1 | 0.04 | 0.86 | 0.07 | |
| $C_{DSL}$ | 0 | 1.00 | 0.42 | 0.60 | 0.66 |
| | 1 | 0.03 | 0.90 | 0.07 | |
| $C_{FI}$ | 0 | 0.98 | 0.47 | 0.64 | 0.55 |
| | 1 | 0.03 | 0.62 | 0.05 | |
| $C_{one}$ | 0 | 1.00 | 0.09 | 0.17 | 0.54 |
| | 1 | 0.02 | 1.00 | 0.05 | |
| $C_{vote}$ | 0 | 0.99 | 0.83 | 0.90 | 0.65 |
| | 1 | 0.06 | 0.48 | 0.11 | |
| $C_{all}$ | 0 | 0.98 | 0.82 | 0.90 | 0.58 |
| | 1 | 0.04 | 0.33 | 0.07 | |
| $B_1$ | 0 | 0.98 | 0.98 | 0.98 | 0.51 |
| | 1 | 0.06 | 0.05 | 0.05 | |
| $B_2$ | 0 | 0.98 | 1.00 | 0.99 | 0.50 |
| | 1 | 0.00 | 0.00 | 0.00 | |
| $B_3$ | 0 | 0.00 | 0.00 | 0.00 | 0.50 |
| | 1 | 0.02 | 1.00 | 0.04 | |
| $B_4$ | 0 | 0.98 | 0.77 | 0.86 | 0.60 |
| | 1 | 0.04 | 0.43 | 0.07 | |

**S3 Table. Injury prediction report of ACWRq.** We report precision (prec), recall (rec), F1-score (F1) and Area Under the Curve (AUC) for the injury class and the non-injury class for all the predictors defined on ACWR and monotony methodologies. We also provide predictive performance of four baseline predictors $B_1$, $B_2$, $B_3$ and $B_4$.

| $ACWR_q$ | class | prec | rec | F1 | AUC |
|---|---|---|---|---|---|
| $C_{dTOT}$ | 0 | 0.98 | 0.80 | 0.88 | 0.59 |
|  | 1 | 0.04 | 0.38 | 0.08 |  |
| $C_{dHSR}$ | 0 | 0.99 | 0.37 | 0.54 | 0.57 |
|  | 1 | 0.03 | 0.76 | 0.05 |  |
| $C_{dMET}$ | 0 | 0.98 | 0.80 | 0.88 | 0.59 |
|  | 1 | 0.04 | 0.38 | 0.08 |  |
| $C_{dHML}$ | 0 | 0.99 | 0.81 | 0.89 | 0.67 |
|  | 1 | 0.06 | 0.52 | 0.10 |  |
| $C_{dHML/m}$ | 0 | 0.98 | 0.81 | 0.89 | 0.62 |
|  | 1 | 0.05 | 0.43 | 0.09 |  |
| $C_{dEXP}$ | 0 | 0.98 | 0.80 | 0.88 | 0.57 |
|  | 1 | 0.04 | 0.33 | 0.07 |  |
| $C_{Acc2}$ | 0 | 0.98 | 0.80 | 0.88 | 0.59 |
|  | 1 | 0.04 | 0.38 | 0.08 |  |
| $C_{Acc3}$ | 0 | 0.98 | 0.80 | 0.64 | 0.54 |
|  | 1 | 0.03 | 0.29 | 0.06 |  |
| $C_{Dec2}$ | 0 | 0.98 | 0.80 | 0.88 | 0.57 |
|  | 1 | 0.04 | 0.33 | 0.07 |  |
| $C_{Dec3}$ | 0 | 0.99 | 0.46 | 0.63 | 0.66 |
|  | 1 | 0.04 | 0.86 | 0.07 |  |
| $C_{DSL}$ | 0 | 0.98 | 0.80 | 0.88 | 0.59 |
|  | 1 | 0.04 | 0.38 | 0.08 |  |
| $C_{FI}$ | 0 | 0.98 | 0.80 | 0.88 | 0.57 |
|  | 1 | 0.04 | 0.33 | 0.07 |  |
| $C_{one}$ | 0 | 0.99 | 0.21 | 0.34 | 0.56 |
|  | 1 | 0.02 | 0.91 | 0.05 |  |
| $C_{vote}$ | 0 | 0.99 | 0.83 | 0.9 | 0.64 |
|  | 1 | 0.06 | 0.46 | 0.1 |  |
| $C_{all}$ | 0 | 0.98 | 1.00 | 0.99 | 0.50 |
|  | 1 | 0.00 | 0.00 | 0.00 |  |
| $B_1$ | 0 | 0.98 | 0.98 | 0.98 | 0.51 |
|  | 1 | 0.06 | 0.05 | 0.05 |  |
| $B_2$ | 0 | 0.98 | 1.00 | 0.99 | 0.50 |
|  | 1 | 0.00 | 0.00 | 0.00 |  |
| $B_3$ | 0 | 0.00 | 0.00 | 0.00 | 0.50 |
|  | 1 | 0.02 | 1.00 | 0.04 |  |
| $B_4$ | 0 | 0.98 | 0.77 | 0.86 | 0.60 |
|  | 1 | 0.04 | 0.43 | 0.07 |  |

**S4 Table. Performance of MSWR predictor.** We report precision (prec), recall (rec), F1-score (F1) and Area Under the Curve (AUC) for the injury class and the non- injury class for all the predictors based on ACWR and MSWR. We also provide predictive performance of four baseline predictors $B_1$, $B_2$, $B_3$ and $B_4$.

| MSWR | class | prec | rec | F1 | AUC |
|---|---|---|---|---|---|
| $C_{dTOT}$ | 0 | 0.98 | 0.80 | 0.88 | 0.57 |
|  | 1 | 0.04 | 0.33 | 0.07 |  |
| $C_{dHSR}$ | 0 | 0.98 | 1.00 | 0.99 | 0.50 |
|  | 1 | 0.00 | 0.00 | 0.00 |  |
| $C_{dMET}$ | 0 | 0.98 | 0.95 | 0.96 | 0.55 |
|  | 1 | 0.06 | 0.14 | 0.09 |  |
| $C_{dHML}$ | 0 | 0.98 | 0.96 | 0.97 | 0.53 |
|  | 1 | 0.06 | 0.10 | 0.07 |  |
| $C_{dHML/m}$ | 0 | 0.98 | 0.96 | 0.97 | 0.55 |
|  | 1 | 0.08 | 0.14 | 0.10 |  |
| $C_{dEXP}$ | 0 | 0.98 | 0.94 | 0.96 | 0.49 |
|  | 1 | 0.02 | 0.05 | 0.03 |  |
| $C_{Acc2}$ | 0 | 0.98 | 0.93 | 0.95 | 0.46 |
|  | 1 | 0.00 | 0.00 | 0.00 |  |
| $C_{Acc3}$ | 0 | 0.98 | 0.98 | 0.98 | 0.49 |
|  | 1 | 0.00 | 0.00 | 0.00 |  |
| $C_{Dec2}$ | 0 | 0.98 | 0.94 | 0.96 | 0.52 |
|  | 1 | 0.04 | 0.10 | 0.05 |  |
| $C_{Dec3}$ | 0 | 0.98 | 0.99 | 0.98 | 0.49 |
|  | 1 | 0.00 | 0.00 | 0.00 |  |
| $C_{DSL}$ | 0 | 0.98 | 0.97 | 0.97 | 0.48 |
|  | 1 | 0.00 | 0.00 | 0.00 |  |
| $C_{FI}$ | 0 | 0.98 | 0.72 | 0.83 | 0.50 |
|  | 1 | 0.03 | 0.29 | 0.04 |  |
| $C_{one}$ | 0 | 0.98 | 0.56 | 0.71 | 0.54 |
|  | 1 | 0.03 | 0.52 | 0.05 |  |
| $C_{vote}$ | 0 | 0.97 | 0.99 | 0.98 | 0.49 |
|  | 1 | 0.00 | 0.00 | 0.00 |  |
| $C_{all}$ | 0 | 0.97 | 1.00 | 0.99 | 0.50 |
|  | 1 | 0.00 | 0.00 | 0.00 |  |
| $B_1$ | 0 | 0.98 | 0.98 | 0.98 | 0.51 |
|  | 1 | 0.06 | 0.05 | 0.05 |  |
| $B_2$ | 0 | 0.98 | 1.00 | 0.99 | 0.50 |
|  | 1 | 0.00 | 0.00 | 0.00 |  |
| $B_3$ | 0 | 0.00 | 0.00 | 0.00 | 0.50 |
|  | 1 | 0.02 | 1.00 | 0.04 |  |
| $B_4$ | 0 | 0.98 | 0.77 | 0.86 | 0.60 |
|  | 1 | 0.04 | 0.43 | 0.07 |  |

**S5 Table.** PI$^{(WF)}$ values after n training days (i.e., $n = 1, \ldots, 6$) since the return of a player to regular training. We report the values for different n of previous injuries (i.e., $n = 1, \ldots, 4$). PI$_i$ is the number of training days long after players return to regular physical activity. 6+ indicates values for 6 and more than 6 days.

|  | PI$_i$ | | | | | |
| --- | --- | --- | --- | --- | --- | --- |
| injuries | 1 | 2 | 3 | 4 | 5 | 6+ |
| 1 | 0.29 | 0.49 | 0.64 | 0.74 | 0.81 | > 0.86 |
| 2 | 1.27 | 1.48 | 1.63 | 1.74 | 1.81 | > 1.86 |
| 3 | 2.27 | 2.46 | 2.62 | 2.72 | 2.8 | > 2.85 |
| 4 | 3.25 | 3.46 | 3.53 | 3.66 | 3.76 | > 3.83 |

**S6 Table Performance of the classifiers on T^(ADA), T and T^(REF).** For each classifier, we report the precision (prec), recall (rec) and F1-score (F1) on the two classes and the overall AUC.

|   |    | \multicolumn{4}{c}{T^(ADA)} | | | | \multicolumn{4}{c}{T} | | | | \multicolumn{4}{c}{T^(RFE)} | | | |
|---|----|------|------|------|------|------|------|------|------|------|------|------|------|
|   |    | *prec* | *rec* | *F1* | *AUC* | *prec* | *rec* | *F1* | *AUC* | *prec* | *rec* | *F1* | *AUC* |
| DT | NI | 0.92 | 0.87 | 0.90 | 0.73 | 0.99 | 0.99 | 0.99 | 0.68 | 0.98 | 1.00 | 0.99 | 0.71 |
|    | I  | **0.57** | **0.72** | **0.64** |      | **0.42** | **0.66** | **0.58** |      | **0.74** | **0.56** | **0.64** |      |
| RF | NI | 0.93 | 0.91 | 0.92 | 0.75 | 0.99 | 1.00 | 0.99 | 0.70 | 0.99 | 0.99 | 0.99 | 0.73 |
|    | I  | 0.71 | 0.63 | 0.70 |      | 0.38 | 0.72 | 0.60 |      | 0.78 | 0.58 | 0.66 |      |
| LR | NI | 0.83 | 0.77 | 0.80 | 0.71 | 0.98 | 0.99 | 0.99 | 0.61 | 0.98 | 0.98 | 0.98 | 0.63 |
|    | I  | 0.68 | 0.64 | 0.65 |      | 0.58 | 0.33 | 0.42 |      | 0.73 | 0.48 | 0.55 |      |

**S7 Table. Feature Selection real-world scenario.** Features extracted by RFECV in each $T_i$ built as the season went by.

| $T_i$ | RFECV |
|---|---|
| 6 | $d_{MET}^{(EWMA)}$, $DEC_3^{(ACWR)}$, $PI^{(EWMA)}$ |
| 7 | $PI^{(EWMA)}$, $d_{HSR}^{(EWMA)}$, $d_{TOT}^{(MSWR)}$ |
| 8 | $PI^{(EWMA)}$, $d_{HSR}^{(EWMA)}$, $d_{TOT}^{(MSWR)}$ |
| 9 | $d_{HSR}^{(EWMA)}$, $ACC_2^{(EWMA)}$, $d_{HML/m}^{(ACWR)}$, $d_{EXP}^{(MSWR)}$, $PI^{(EWMA)}$ |
| 10 | $d_{HSR}^{(EWMA)}$, $PI^{(EWMA)}$ |
| 11 | $DEC_2$, $d_{HSR}^{(EWMA)}$, $d_{HML/m}^{(EWMA)}$, $d_{EXP}^{(MSWR)}$, $PI^{(EWMA)}$ |
| 12 | $ACC_2$, $d_{HSR}^{(EWMA)}$, $DEC_3^{(ACWR)}$, $PI^{(EWMA)}$ |
| 13 | $d_{HSR}^{(EWMA)}$, $d_{HSR}^{(ACWR)}$, $PI^{(EWMA)}$, $FI^{(MSWR)}$ |
| 14 | $PI^{(EWMA)}$, $d_{HSR}^{(EWMA)}$, $d_{TOT}^{(MSWR)}$ |
| 15 | $ACC_2^{(EWMA)}$, $PI^{(EWMA)}$ |
| 16 | $PI^{(EWMA)}$, $d_{HSR}^{(EWMA)}$, $d_{TOT}^{(MSWR)}$ |
| 17 | $PI^{(EWMA)}$, $d_{HSR}^{(EWMA)}$, $d_{TOT}^{(MSWR)}$ |
| 18 | $PI^{(EWMA)}$, $d_{HSR}^{(EWMA)}$, $d_{TOT}^{(MSWR)}$ |
| 19 | $PI^{(EWMA)}$, $d_{HSR}^{(EWMA)}$, $d_{TOT}^{(MSWR)}$ |
| 20 | $PI^{(EWMA)}$, $d_{HSR}^{(EWMA)}$, $d_{TOT}^{(MSWR)}$ |
| 21 | $PI^{(EWMA)}$, $d_{HSR}^{(EWMA)}$, $d_{TOT}^{(MSWR)}$ |

**S1 Fig. Distribution of workload features.** We provide three categories of training workload features: kinematic features (blue), metabolic features (red) and mechanical features (green).

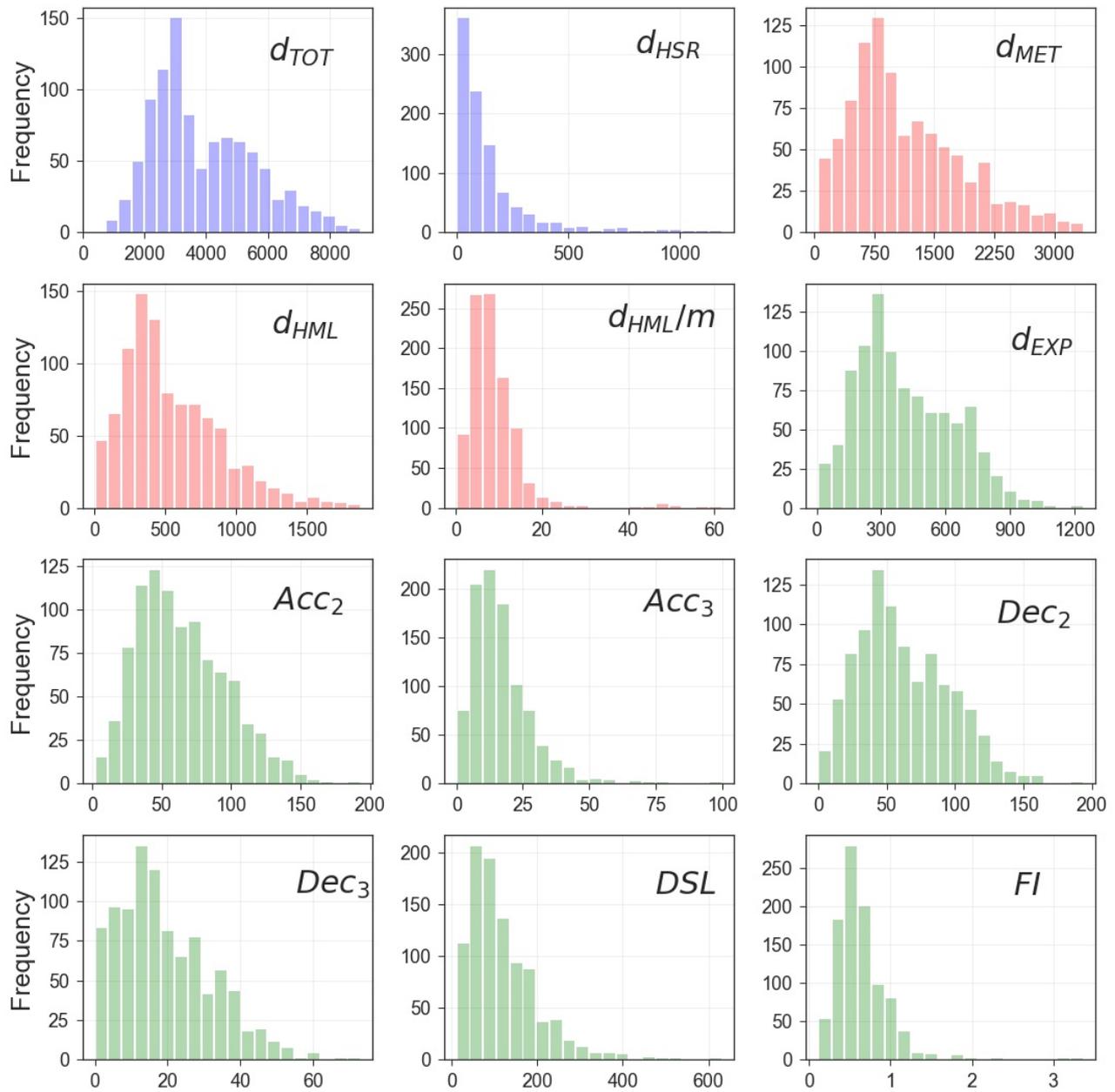

**S2 Fig. Injury risk in ACWR groups.** The plots show Injury Likelihood (IL) for pre-defined ACWR groups [29], for every of the 12 training workload features considered in our study. Bars are colored according to feature categorization defined in Table 1.

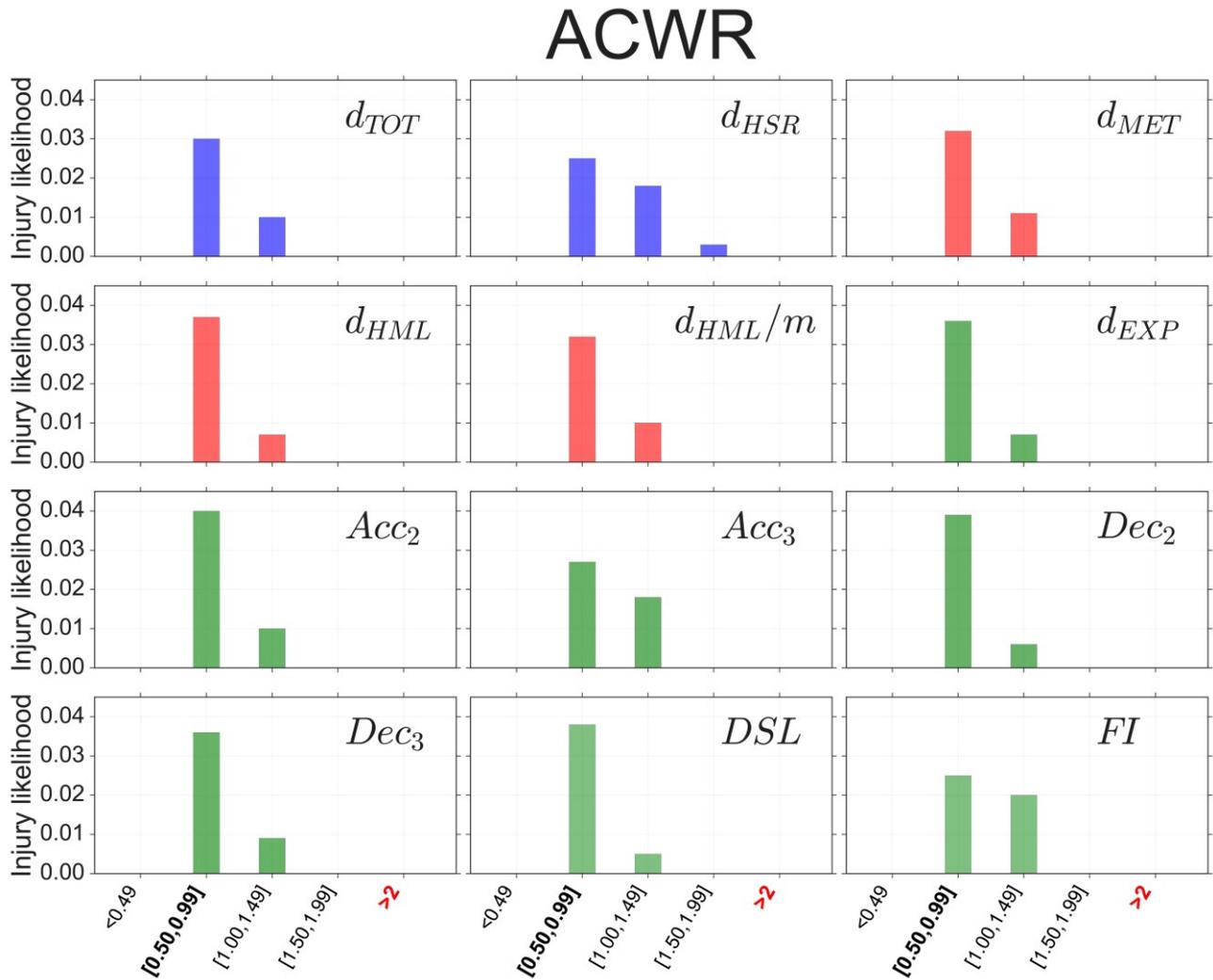

**S3 Fig. Injury likelihood in ACWR groups.** The plots show IL for the ACWR groups defined the quantiles of the distribution, for every of the 12 training workload features considered in our study. We provide three categories of training workload features: kinematic features (blue), metabolic features (red) and mechanical features (green).

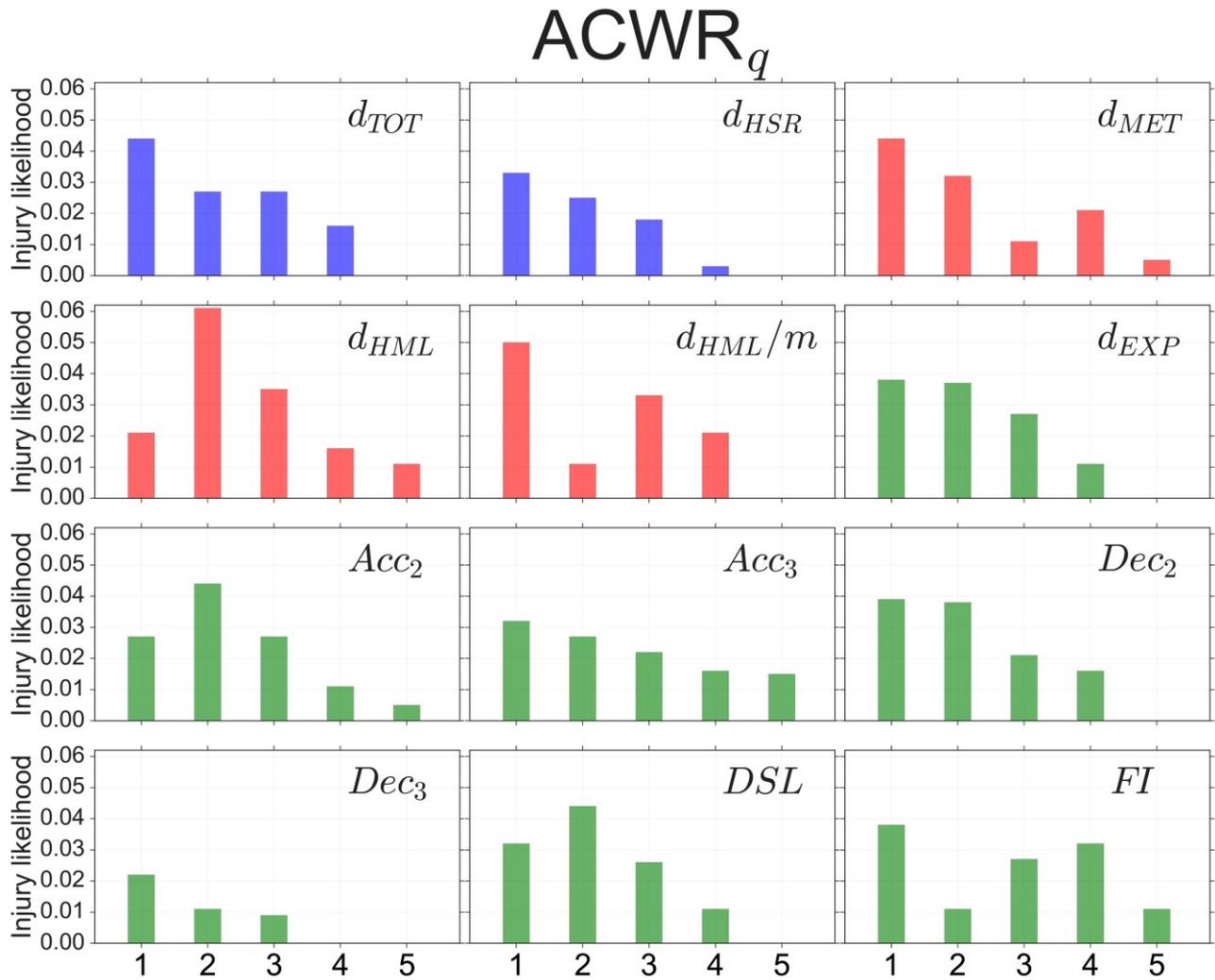

**S4 Fig. Injury risk in MSWR groups.** The plots show the Injury Likelihood (IL) for the MSWR groups for every of the 12 training workload features considered in our study. Bars are colored according to feature categorization defined in Table 1.

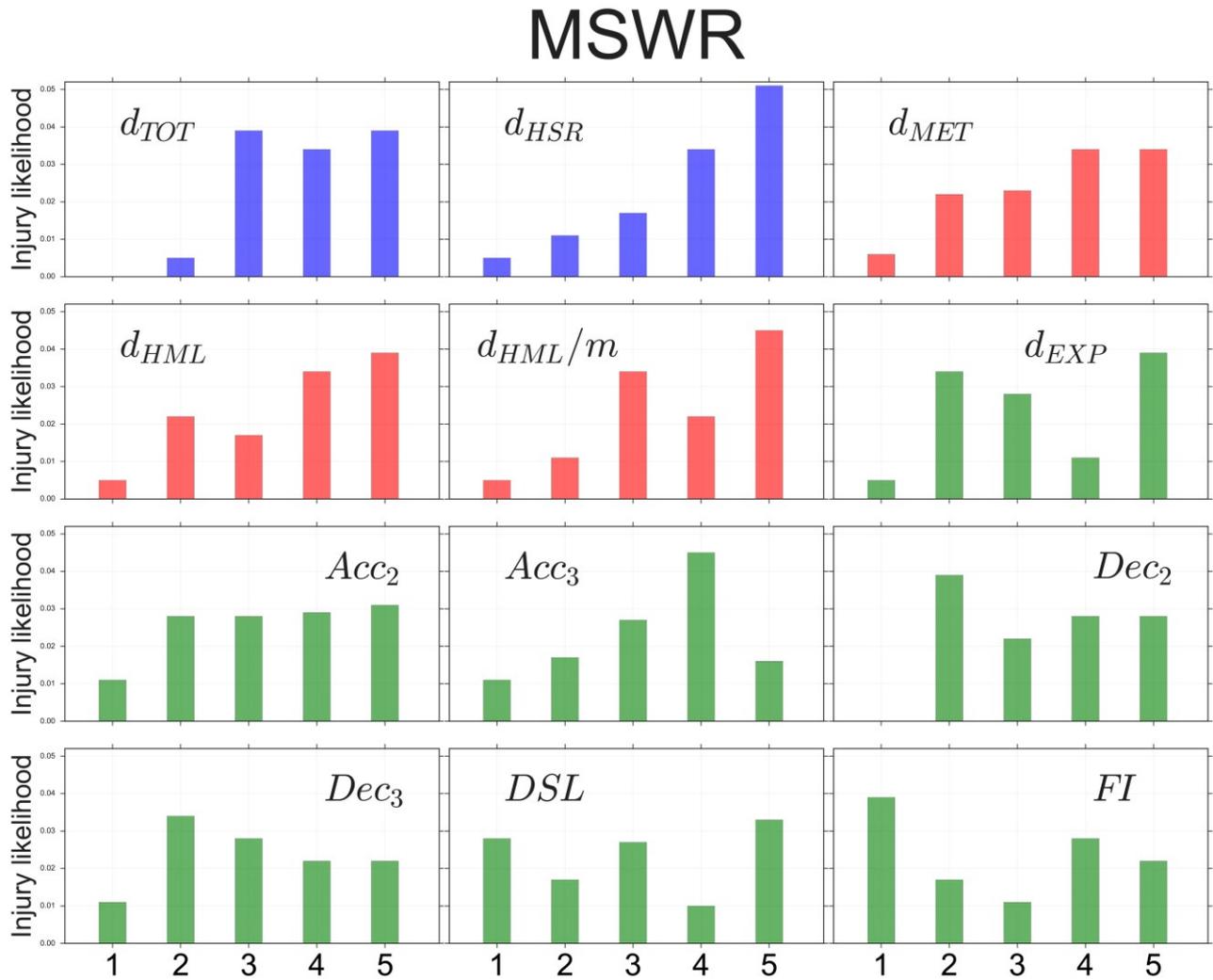

**S5 Fig.** We plot the AUC and F1-score of EWMA with span = 1, . . . , 10 in CALL. The red line reflects the best span to injury prediction.

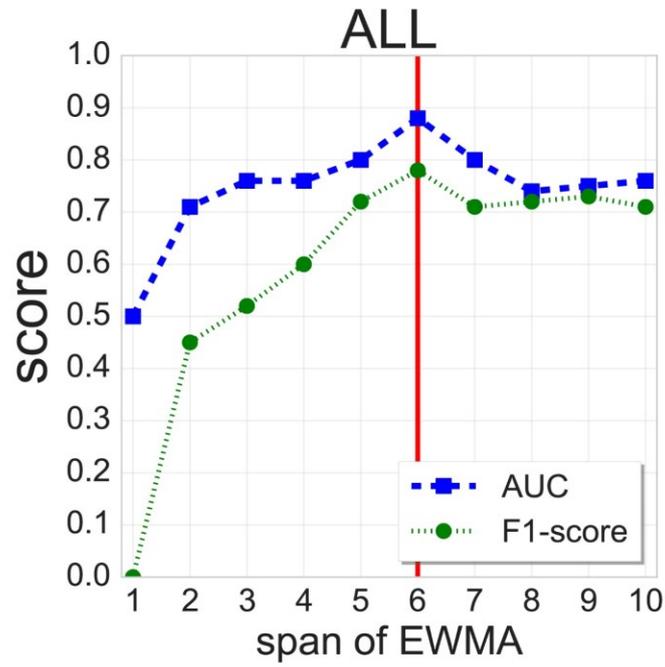